# Deep Imitation Learning for Complex Manipulation Tasks from Virtual Reality Teleoperation

Tianhao Zhang[*12], Zoe McCarthy[*1], Owen Jow[1], Dennis Lee[1], Xi Chen[12], Ken Goldberg[1], Pieter Abbeel[1-4]

*Abstract*— Imitation learning is a powerful paradigm for robot skill acquisition. However, obtaining demonstrations suitable for learning a policy that maps from raw pixels to actions can be challenging. In this paper we describe how consumer-grade Virtual Reality headsets and hand tracking hardware can be used to naturally teleoperate robots to perform complex tasks. We also describe how imitation learning can learn deep neural network policies (mapping from pixels to actions) that can acquire the demonstrated skills. Our experiments showcase the effectiveness of our approach for learning visuomotor skills.

## I. INTRODUCTION

Imitation learning is a class of methods for acquiring skills by observing demonstrations (see, e.g., [1], [2], [3] for surveys). It has been applied successfully to a wide range of domains in robotics, for example to autonomous driving [4], [5], [6], autonomous helicopter flight [7], gesturing [8], and manipulation [9], [10].

High-quality demonstrations are required for imitation learning to succeed. It is straightforward to obtain human demonstrations for car driving [4], [5] or RC helicopter flight [7], because existing control interfaces allow human operators to perform sophisticated, high-quality maneuvers in these domains easily. By contrast, it has been challenging to collect high-quality demonstrations for robotic manipulation. Kinesthetic teaching, in which the human operator guides the robot by force on the robot's body, can be used to gather demonstrations [9], [11], but is unsuitable for learning visuomotor policies that map from pixels to actions due to the unwanted appearance of human arms.

Demonstrations could instead be collected from running trajectory optimization [12], [13], [14], [15] or reinforcement learning [16], [17], [18], [19], [20], [21], [22], but these methods require well-shaped, carefully designed reward functions, access to dynamics model, and substantial robot interaction time. Since these requirements are challenging to meet even for robotics experts, generating high-quality demonstrations programmatically for a wide range of manipulation tasks remains impractical for most situations.

Teleoperation systems designed for robotic manipulation, such as the da Vinci Surgical System developed by Intuitive Surgical Inc. [23], allow high-quality demonstrations to be easily collected without any visual obstructions. Such systems,

*These authors contributed equally to this work.
[1]Department of Electrical Engineering and Computer Science, University of California, Berkeley
[2]Embodied Intelligence
[3]OpenAI
[4]International Computer Science Institute (ICSI)

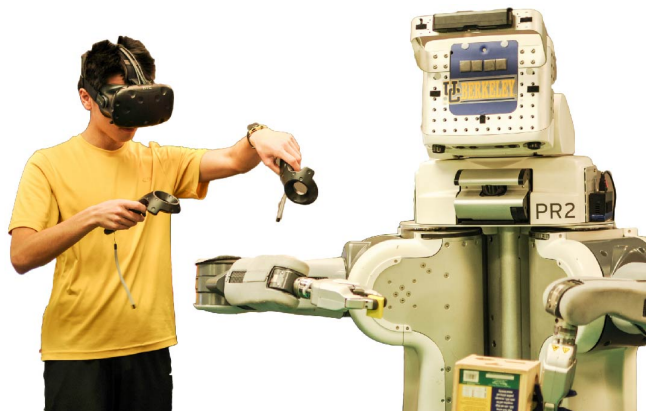

Fig. 1: Virtual Reality teleoperation in action

however, can be expensive and are oftentimes tailored towards specialized hardware.

So we set out to answer two questions in this paper:
- Can we build an inexpensive teleoperation system that allows intuitive robotic manipulation and collection of high-quality demonstrations suitable for learning?
- With high-quality demonstrations, can imitation learning succeed in solving a wide range of challenging manipulation tasks using a practical amount of data?

To answer the first question, we built a system that uses consumer-grade Virtual Reality (VR) devices to teleoperate a PR2 robot. A human operator of our system uses a VR headset to perceive the environment through the robot's sensor space, and controls the robot with motion-tracked VR controllers in a way that leverages the natural manipulation instincts that humans possess (see Fig. 1). This setup ensures that the human and the robot share exactly the same observation and action space, eliminating the possibility of the human making any decision based on information not available to the robot, and preventing visual distractions, like human hands for kinesthetic teaching, from entering the environment.

To answer the second question, we collected demonstrations using our system on ten real-world manipulation tasks with a PR2 robot, and trained deep visuomotor policies that directly map from pixels to actions using behavioral cloning, a simple imitation learning method. We show that behavioral cloning, with high-quality demonstrations, is surprisingly effective. For an expanded description of each task, please see our supplemental video and supplemental website[1].

In summary, our main contributions are:

[1]https://sites.google.com/view/vrlfd

- We built a VR teleoperation system on a real PR2 robot using consumer-grade VR devices.
- We proposed a single neural network architecture (Fig. 3) for all tasks that maps from raw color and depth pixels to actions, augmented with auxiliary prediction connections to accelerate learning.
- Perhaps our most surprising finding is that, for each task, less than 30 minutes of demonstration data is sufficient to learn a successful policy, with the same hyperparameters and neural network architecture used across all tasks.

## II. Related Work

Two main lines of work within imitation learning are behavioral cloning, which performs supervised learning from observations to actions (e.g., [4], [24]) and inverse reinforcement learning [25], where a reward function [26], [27], [28], [29], [30] is estimated to explain the demonstrations as (near) optimal behavior. This work focuses on behavioral cloning.

Behavioral cloning has led to many successes in robotics when a low-dimensional representation of environment state is available [31], [32], [33], [34]. However, it is often challenging to extract state information and hence more desirable to learn policies that directly take in raw pixels. This approach has proven successful in domains where collecting such demonstrations is natural, such as simulated environments [24], [35], driving [4], [5], [6], and drones [36].

For real-world robotic manipulation, however, collecting demonstrations suitable for learning visual policies is difficult. Kinesthetic teaching is not intuitive and can result in unwanted visual artifacts [9], [11]. Using motion capture devices for teleoperation, such as [37], is more intuitive and can solve this issue. However, the human teacher typically observes the scene through a different angle from the robot, which may render certain objects only visible to the human or the robot (due to occlusions), making imitation challenging. Another approach is to collect third-person demonstrations, such as raw videos [38], [39], but this poses challenges in learning.

On the other hand, Virtual Reality teleoperation allows for a direct mapping of observations and actions between the teacher and the robot and does not suffer from the above correspondence issues [3], while also leveraging the natural manipulation instincts that the human teacher possesses. In a non-learning setting, VR teleoperation has been recently explored for controlling humanoid robots [40], [41], [42], for simulated dexterous manipulation [43], and for communicating motion intent [44]. Existing use cases of VR for learning policies have so far been limited to collecting waypoints of low-dimensional robot states [45], [46].

Reinforcement learning (RL) provides an alternative for skill acquisition, where a robot acquires skills from its own trial and error. While more traditional RL success stories in robotics (e.g., [47], [48], [49]) work in state space, more recent work has been able to learn deep neural net policies from pixels to actions (e.g., [17], [18], [20]). While learning policies from pixels to actions has been remarkably successful, the amount of exploration required can often be impractical for real robot systems (e.g., the Atari results would have taken

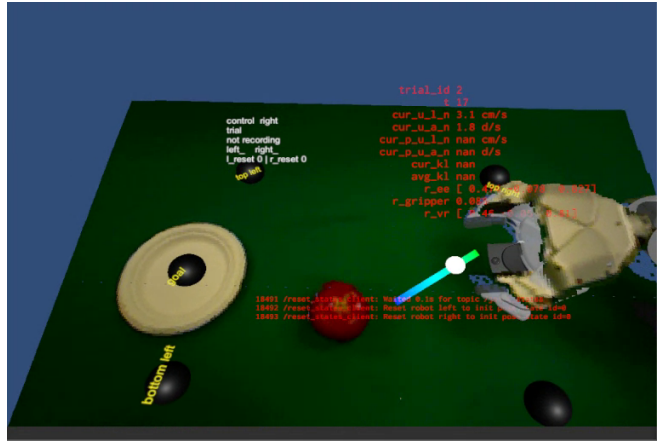

Fig. 2: First-person view from inside our VR teleoperation system during a demonstration, which includes VR visualizations helpful for the human operator (see Section III-B).

40 days of real-time experience). Guided Policy Search [20] is able to significantly reduce sample complexity, but in turn it relies on using trajectory-centric reinforcement learning to autonomously obtain demonstrations of the task at hand. In addition, reinforcement learning algorithms require a reward function, which can be difficult to specify in practice [50].

## III. Virtual Reality Teleoperation

In this section, we describe our Virtual Reality teleoperation system and discuss how it allows humans to naturally produce demonstrations suitable for learning policies from pixels.

### A. Hardware

We base our teleoperation platform on the Vive VR system, a consumer-grade VR device that costs \$600, and a PR2 robot. The Vive provides a headset for head-mounted display and two hand controllers, each with 6 DoF pose tracking at sub-millimeter precision at 90 Hz in a room-scale tracking area. For visual sensing, we use the Primesense Carmine 1.09, a low-cost 3D camera, mounted on the robot head for providing first-person color and depth images at 30 Hz. Our teleoperation system is written in Unity, a 3D game engine that supports major VR headsets such as the Vive.

### B. Visual Interface

We designed the visual environment of our teleoperation system to be informative and intuitive for human operators, to best leverage their intuition about 3D space during teleoperation, while remaining comfortable to operate for extended periods of time.

One may imagine presenting the scene to the user by displaying a pair of images, captured from a stereo camera, into the two lenses of the VR head-mounted display. While easy to implement, this scheme can lead to motion sickness, since it is difficult to ensure the displayed scene is consistent with the human operator's head motion with little time lag. The robot head, where the camera is mounted, is not only less precise and agile compared to the human head, but also has fewer degrees of freedom: the human head can move with a full six degrees of freedom, but the PR2's head has two. To compensate, the PR2's slow base and torso would have to move in addition to achieve a given 6D head pose, leading to greater potential inconsistency and lag between the user's head pose movements and the displayed scene.

To avoid these problems, we use an RGB-D camera to capture color images with per-pixel depth values, and we render the corresponding colored 3D point cloud, processed to remove gaps between points, as physical objects in the virtual environment. The human operator views the environment via a virtual camera whose pose is instantaneously updated to reflect the operator's head movement. This allows us to avoid motion sickness. Note that allowing head movements does not change the information available to the operator. In addition, this approach allows useful 3D visualizations to be overlayed on the point cloud to assist the operator throughout the teleoperation process. For example, markers can be placed at specified 3D locations to instruct operators where to initialize objects during training, and a 3D arrow indicating intended human control can be plotted alongside other textual displays. See Fig. 2 and supplemental video for views from within our system.

### C. Control Interface

We use the Vive hand controllers for controlling the robot's arms and use the trigger button on the controller to signal the robot gripper to fully open or close. Thanks to the immersive visual interface made possible by VR, we can map the human operator and the robot to a unified coordinate frame in the virtual environment, where the pose of the operator's hand, tracked by the VR controller, is interpreted as the target pose of the robot's corresponding gripper. We collect the target pose of the gripper at 10 Hz, which is used by a built-in low-level Jacobian-transpose based controller to control the robot arm at 1k Hz at the torque level.

This control mechanism is very natural, because humans can simply move their hand and the pose target for the robot gripper is moved in the same way, making it easy for even first-time users to accomplish complex manipulation tasks. There is minimal difference in kinematics in this setting – unlike kinesthetic teaching, where the human operators must use very different movement than they would naturally to achieve the same motion of their hands. In addition, the operator receives instantaneous feedback from the environment, such as how objects in the environment react to the robot's movements. Another advantage of this control scheme is that it provides an intuitive way to apply force control. When the gripper is not obstructed by any object, the low-level controller effectively performs position control for the gripper. However, when the gripper starts making contact and becomes hindered, which often happens in the contact-rich manipulation tasks considered in this paper, the magnitude of the difference between the target pose and the instantaneous pose will scale proportionally with the amount of force exerted by the gripper. This allows the human operator to dynamically vary the force as needed, for example, during insertion and pushing, after visually observing discrepancies between the actual and desired gripper poses.

## IV. LEARNING

Here we present a simple behavioral cloning algorithm to learn our neural network control policies. This entails collecting a dataset $\mathcal{D}_{task} = \{(o_t^{(i)}, u_t^{(i)})\}$ that consists of example pairs of observation and corresponding controls through multiple demonstrations for a given task. The neural network policy $\pi_\theta(u_t|o_t)$, parametrized by $\theta$, then learns a function that reconstructs the controls from the observation for each example pair.

### A. Neural Network Control Policies

The inputs $o_t = (I_t, D_t, p_{t-4:t})$ at time step $t$ to the neural network policy includes (a) current RGB image $I_t \in \mathbb{R}^{160 \times 120 \times 3}$, (b) current depth image $D_t \in \mathbb{R}^{160 \times 120}$ (both collected by the on-board 3D camera), and (c) three points on the end effector of the right arm, used for representing pose similar to [20], for the 5 most recent steps $p_{t-4:t} \in \mathbb{R}^{45}$.

Including the short history of the end-effector points allows the robot to infer velocity and acceleration from the kinematic state. We choose not to include the 7 dimensional joint angles of the right arm as inputs since the human operator can only directly control the position and orientation, which collectively are the 6 DoF of the end effector.

The neural network outputs the current control $u_t$, which consists of angular velocity $\omega_t \in \mathbb{R}^3$ and linear velocity $v_t \in \mathbb{R}^3$ of the right hand, as well as the desired gripper open/close $g_t \in \{0, 1\}$ for tasks involving grasping. Although our platform supports controlling both arms and the head, for simplicity we only subjected the right arm to control and froze all other joints except when resetting to initial states.[2] During execution, the policy $\pi_\theta$ generates the control $u_t = \pi_\theta(o_t)$ given current observation $o_t$. Observations and controls are both collected at 10 Hz.

Our neural network architecture, as shown in Fig. 3, closely follows [20], except that we additionally provide depth image as input and include auxiliary prediction tasks to accelerate learning. Concretely, our neural network policy $\pi_\theta$ can be decomposed into three modules $\theta = (\theta_{vision}, \theta_{aux}, \theta_{control})$. Given observation $o_t$, a convolutional neural network with a spatial soft-argmax layer [20] first extracts spatial feature points from images (Eq. 1), followed by a small fully-connected network for auxiliary prediction (Eq. 2), and finally another fully-connected network outputs the control (Eq. 3). Except for the final layer and the spatial soft-argmax layer, each layer is followed by a layer of rectified linear units.

$$f_t = \text{CNN}(I_t, D_t; \theta_{vision}) \quad (1)$$
$$s_t = \text{NN}(f_t; \theta_{aux}) \quad (2)$$
$$u_t = \text{NN}(p_{t-4:t}, f_t, s_t; \theta_{control}) \quad (3)$$

### B. Loss Functions

The loss function used for our experiments is a small modification to the standard loss function for behavioral cloning. Given an example pair $(o_t, u_t)$, behavioral cloning algorithms typically use $l1$ and $l2$ losses to fit the training data:

$$\mathcal{L}_{l2} = ||\pi_\theta(o_t) - u_t||_2^2, \quad \mathcal{L}_{l1} = ||\pi_\theta(o_t) - u_t||_1 \quad (4)$$

---

[2]The left arm may move in face of sufficient external force, such as in the plane task.

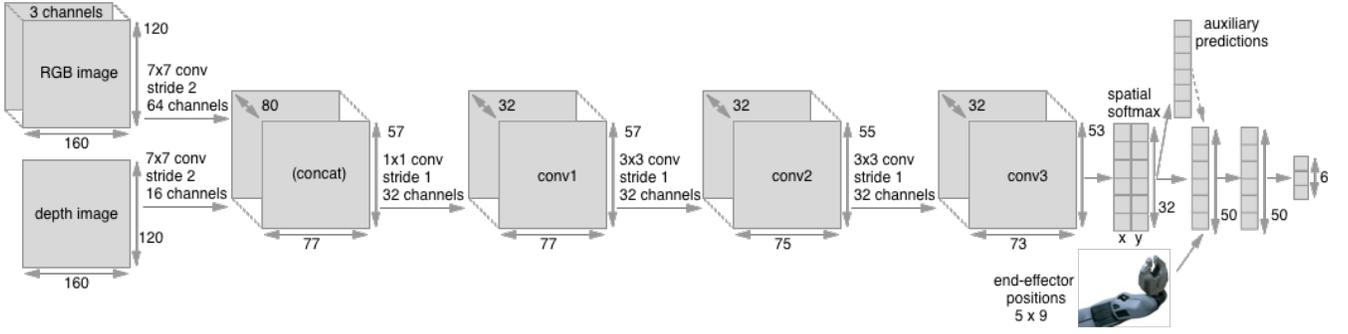

Fig. 3: Architecture of our neural network policies

Since we care more about the direction than the magnitude of the movement of the end effector, we also introduce a loss to encourage directional alignment between the demonstrated controls and network outputs, as follows (note the $\arccos$ outputs are in the range of $[0, \pi]$):

$$\mathcal{L}_c = \arccos\left(\frac{u_t^T \pi_\theta(o_t)}{||u_t||\, ||\pi_\theta(o_t)||}\right) \quad (5)$$

For tasks that involve grasping, the final layer outputs a scalar logit $\hat{g}_t$ for gripper open/close prediction $g_t \in \{0, 1\}$, which we train using sigmoid cross entropy loss:

$$\mathcal{L}_g = g_t \log(\sigma(\hat{g}_t)) - (1 - g_t)\log(1 - \sigma(\hat{g}_t)) \quad (6)$$

The overall loss function is a weighted combination of standard loss functions, as described above, and additional loss functions for auxiliary prediction tasks (see Section IV-C).

$$\mathcal{L}(\theta) = \lambda_{l2}\mathcal{L}_{l2} + \lambda_{l1}\mathcal{L}_{l1} + \lambda_c\mathcal{L}_c + \lambda_g\mathcal{L}_g + \lambda_{aux}\sum_a \mathcal{L}_{aux}^{(a)} \quad (7)$$

We used stochastic gradient descent to train our neural network policies with batches randomly sampled from $\mathcal{D}_{task}$.

### C. Auxiliary Loss Functions

We include auxiliary prediction tasks as an extra source of self-supervision. Similar approaches that leverage self-supervisory signals were shown by [51] to improve data efficiency and robustness. For each auxiliary task $a$, a small module of two fully-connected layers is added after the spatial soft-argmax layer, i.e. $\hat{s}_t^{(a)} = \text{NN}(f_t; \theta_{aux}^{(a)})$, and is trained using $l2$ loss with label $s_t^{(a)}$:

$$\mathcal{L}_{aux}^{(a)} = ||\text{NN}(f_t; \theta_{aux}^{(a)}) - s_t^{(a)}||_2^2 \quad (8)$$

In our experiments, the labels $s_t^{(a)}$ for these auxiliary tasks can be readily inferred from the dataset $\mathcal{D}_{task}$, such as the current gripper pose $p_t$ and the final gripper pose $p_T$. This resembles the pretraining process in [20], where the CNN is pretrained with a separate dataset of images with labeled gripper and object poses, but our approach requires no additional dataset and all training is done concurrently.

## V. EXPERIMENTS

Our primary goal is to empirically investigate the effectiveness of simple imitation learning using demonstrations collected via Virtual Reality teleoperation:

(i) Can we use our system to train, with little tuning, successful deep visuomotor policies for a range of challenging manipulation tasks?

In addition, we strive to further understand the results by analyzing the following aspects:

(ii) What is the sample complexity for learning an example manipulation task using our system?
(iii) Does our auxiliary prediction loss improve data efficiency for learning real-world robotic manipulation?

In this section, we describe a set of experiments on a real PR2 robot to answer these questions. Our findings are somewhat surprising: while folk wisdom suggests deep learning from raw pixels would require large amounts of data, with under 30 minutes of demonstrations for each task, the learned policies already achieve high success rates and good generalization.

### A. Experimental Setup

We chose a range of challenging manipulation tasks (see Fig. 4), where the robot must (a) reach a bottle, (b) grasp a tool, (c) push a toy block, (d) attach wheels to a toy plane, (e) insert a block onto a shape-sorting cube, (f) align a tool with a nail, (g) grasp and place a toy fruit onto a plate, (h) grasp and drop a toy fruit into a bowl and push the bowl, (i) perform grasp-and-place in sequence for two toy fruits, (j) pick up a piece of disheveled cloth.

Successful control policies must learn object localization (a, b, c, g, h, i), high-precision control (a, f, e), managing simple deformable objects (j), and handling contact (c, d, e, f, h, i), all on top of good generalization. Since imitation learning often suffers from poor long horizon performance due to compounding errors, we added tasks (g, h, i) that require multiple stages of movements and a longer duration to complete. We chose tasks (d, e, f) because they were previously used to demonstrate the performance of state-of-the-art algorithms for real-world robotic manipulation [20]. See Appendix VI-A for detailed task specifications, descriptions of the initial states, and the success metrics used for test-time evaluation.

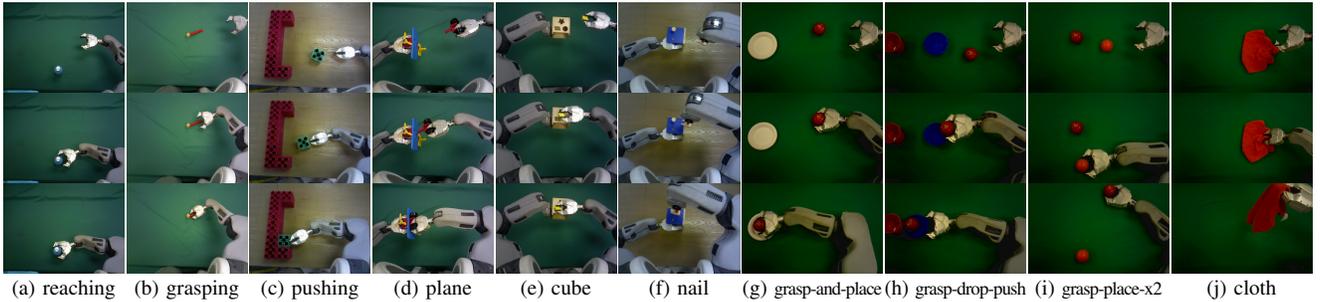

(a) reaching  (b) grasping  (c) pushing  (d) plane  (e) cube  (f) nail  (g) grasp-and-place  (h) grasp-drop-push  (i) grasp-place-x2  (j) cloth

Fig. 4: Examples of successful trials performed by the learned policies during evaluation. Each column shows the image inputs $I_t$ at $t = 0, \frac{T}{2}, T$ for the corresponding task.

TABLE I: Top: success rates of the learned policies averaged across all initial states during test time (see Sec. V-B for details). Bottom: statistics of training data, including total time during demonstration, average length of demonstrations, and total number of demonstrations.

| task | reaching | grasping | pushing | plane | cube | nail | grasp-and-place | grasp-drop-push | grasp-place-x2 | cloth |
|---|---|---|---|---|---|---|---|---|---|---|
| test | 91.6% | 97.2% | 98.9% | 87.5% | 85.7% | 87.5% | 96.0% | 83.3% | 80% | 97.4% |
| demo time (min) | 13.7 | 11.1 | 16.9 | 25.0 | 12.7 | 13.6 | 12.3 | 14.5 | 11.6 | 10.1 |
| avg length (at 10 Hz) | 41 | 37 | 58 | 47 | 37 | 38 | 68 | 87 | 116 | 60 |
| # demo | 200 | 180 | 175 | 319 | 206 | 215 | 109 | 100 | 60 | 100 |

We collected demonstrations for each task using our VR teleoperation system (see Table I for summary). As our goal was to validate the feasibility of our method, we did not perform an explicit search for the minimum number of demonstrations required for learning successful policies. Furthermore, interaction with the robot usually took place in a single session, unlike iterative learning algorithms which require interspersed data collection between iterations.

In addition to having a sufficient number of samples, learning a successful policy also requires sufficient variations in the training data. While prior methods, such as GPS [20], rely on linear Gaussian controllers to inject desired noise, we found that demonstrations collected by human operators naturally display sufficient local variations, as shown in Fig. 5.

### B. Results and Analysis

In the following subsections, we answer the questions we put forth at the beginning of this section.

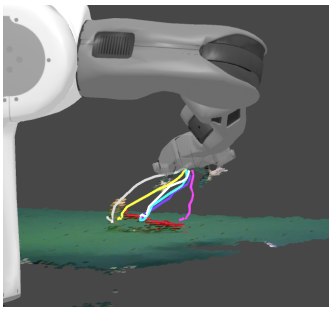

Fig. 5: Overlay of six demonstration trajectories starting from the same initial state for the grasping task.

*Question (i) Can we use our system to train, with little tuning, successful deep visuomotor policies for a range of challenging manipulation tasks?*

To answer this question, we trained a neural network policy for each task using the procedure summarized in the preceding sections. In particular, we used a fixed set of hyperparameters (including neural net architecture) across all the tasks. To explore the effectiveness of this simple learning algorithm, we used a small amount of demonstrations (all under 30 minutes' worth) as training data for each task (see Table I).

We evaluated the learned policies at unseen initial states at test time. Specifications of the initial states and the success metric can be found in Appendix VI-A. While evaluating tasks involving free-form objects (i.e. all except d, e, f), we selected object initial positions uniformly distributed within the training regime, with random local variations around these positions.

Table I shows the success rates of our learned policies for all tasks, while Fig. 4 depicts illustrations of successful trials performed by the learned policies. Surprisingly, with under 30 minutes of demonstrations for each task, all learned policies achieved high success rates and good generalization to test situations. The results suggest that a simple imitation learning algorithm can train successful control policies for a range of real-world manipulation tasks, while achieving tractable sample efficiency and good performance, even in long running tasks.

In addition to successfully completing the tasks, the policies in some cases demonstrated good command of the acquired skills. In the pushing task, the robot learned how to balance the block to maintain the correct direction using a single point of contact (see Fig. 6). In the plane task, the policy chose to wiggle slightly only when movement came to a halt in the middle of the insertion sequence. It is worth noting that the policies were able to complete a sequence of maneuvers in long-running tasks, which shows that the policies also learned how to transition from one skill to another. Fig. 6 showcases a successful trial performed by the learned policy on the grasp-place-x2 task.

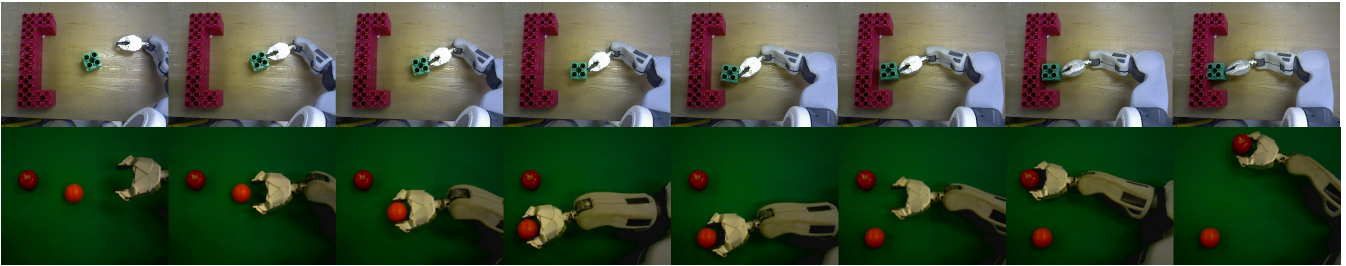

Fig. 6: Example successful trials of the learned policies during evaluation (top: pushing; bottom: grasp-place x2)

*Suboptimal Behaviors:* While achieving success according to our metric, the learned policies were often suboptimal compared to human demonstrations. A common case is that the robot did not follow the shortest path to the goal as in the demonstrations, moved slowly, or paused entirely before resuming motion. In tasks involving grasping, the robot might accidentally nudge the object or attempt grasping several times.

*Failure Cases:* For each task, we report their failure behaviors: (a) knocked over the bottle during reaching, (b) went to the correct position of the tool but failed to close the gripper, (c) stuck the block onto the upper or lower boundaries of the target zone, (d) did not land the peg of the wheels onto the plane, (e) stopped moving when the block grew near (<3 mm) to the slot, and (f) missed the nail and failed to align, (g) failed to grasp the apple or collided with the plate, (h) refused to move after dropping the apple or toppled the bowl during pushing, (i) failed to grasp the second toy fruit, and (j) did not descend far enough to successfully grasp the cloth.

TABLE II: Success rates of policies trained using different numbers of demonstrations for the nail task

| task: nail | | |
| --- | --- | --- |
| number of demonstrations | demonstration time (estimated) (min) | success rates |
| 193 | 12.2 | 88.9% |
| 115 | 7.3 | 77.8% |
| 67 | 4.2 | 50% |

*Extrapolation:* We further evaluated the policies at initial states beyond the training regime to explore the limits of the demonstrated generalization. Qualitatively, in the reaching task, the policy rejected a previously unseen green bottle when it was present along with the training bottle. In the pushing task, the robot succeeded even with the block initialized to a position 10 cm lower than any training state. In the grasping task, the policy could handle a hammer placed 6 cm away from the training regime in any direction. Most notably, the policy for the nail task could generalize to new hammer orientations and positions (see Fig. 7b), as well as an unseen nail position the width of the nail's head (3.5 cm) away from the fixed position used during every demonstration.

*Question (ii) What is the sample complexity for learning an example manipulation task using our system?*

While the learned policies were able to achieve high success rates with a modest amount of demonstrations for all tasks, we are still interested in exploring the boundaries in order to better understand the data efficiency of our method. We chose the nail task and the grasp-and-place task as examples and trained separate policies using progressively smaller subsets of the available demonstrations. We evaluated these policies on the same sets of initial states and report the performance in Table II (nail) and Table III (grasp-and-place). As expected, the performance degrades with smaller amounts of training data, so in principle more demonstrations would further improve on the performances we observed. It is worth noting that only 5 minutes of human demonstrations was needed to achieve 50% success for the nail task.

*Question (iii) Does our auxiliary prediction loss improve data efficiency for learning real-world robotic manipulation?*

Motivated by [51] where auxiliary prediction of self-supervisory signals were shown to improve data efficiency for simulated games, we introduced similar auxiliary losses

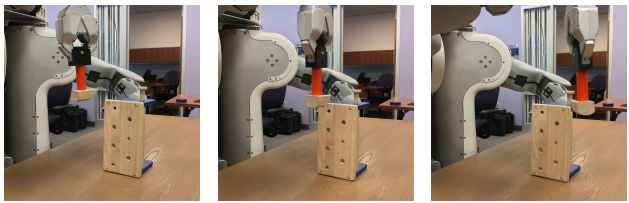

(a) Training: bottom, mid, top (from left to right)

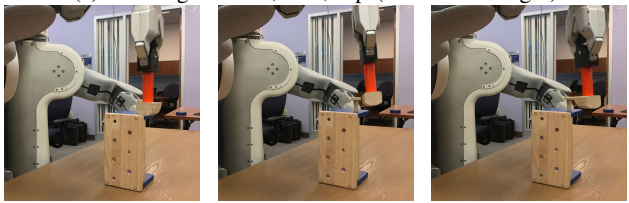

(b) Unseen (extrapolation, see Section V-B)

Fig. 7: Initial states for the nail task

TABLE III: Comparison of policy when trained with and without auxiliary prediction loss on the grasp-and-place task.

| task: grasp-and-place | | |
|---|---|---|
| number of demonstrations | success rates (with) | success rates (without) |
| 109 | 96% | 80% |
| 55 | 53% | 26% |
| 11 | 28% | 20% |

using signals that require no additional efforts to harvest from training demonstrations. It is still interesting to explore whether the same effect can be found for real-world robotic manipulation. We trained policies for the grasp-and-place task with and without auxiliary losses using varying amounts of demonstrations and compared their performance in Table III. We observed that including auxiliary losses indeed empirically improves the data efficiency.

## VI. CONCLUSION

In this paper, we described a VR teleoperation system that makes it easy to collect high-quality robotic manipulation demonstrations that are suitable for visuomotor learning. Then we present the finding that imitation learning can be surprisingly effective in learning deep policies that map directly from pixel values to actions, only with a small amount of learning data. Trained by imitation learning, a single policy architecture from RGB-D images to actions was shown to successfully learn a range of complex manipulation tasks on a real PR2 robot. We empirically studied the generalization of the learned policies and the data efficiency of this learning approach, and show that less than 30 minutes of demonstrations was required to achieve high success rates in novel situations for all the evaluated tasks.

While our current use of VR proved useful for natural demonstration collection, we can further exploit VR to allow for intuitive policy diagnosis through visualization of policy states, collecting additional demonstration signals such as human-provided control variance, and richer feedback to demonstrators such as haptics and sound. On the learning side, since our system allows controlling the head and both arms of the robot, it would be interesting to learn policies with bimanual manipulation or hand-eye coordination. Another exciting direction of future research is scaling up the system to multiple robots for faster, parallel data collection.

## APPENDIX

### A. Task Specification

a) **Reaching**: This task required the robot to reach for a bottle placed at a random position within a 30 x 50 cm accessible region to the left of the right gripper, which was initialized to a fixed pose before reaching. This is challenging because it is easy to knock down the bottle. We consider a trial successful when the bottle can be grasped at the end if the gripper is manually closed.

b) **Grasping**: In this task, the robot must grasp a toy hammer on the table. The hammer was placed randomly in a 15 x 15 cm area at up to $45°$ above or below horizontal and the gripper was initialized as pointing $45°$ upwards, $45°$ downwards or horizontal towards the center. Upon a successful trial, the tool should not be dropped if we manually move the gripper.

c) **Pushing**: This task required the robot to use a closed gripper to push a LEGO block into a fixed target zone. The block may start in a random position and orientation within a 40 x 20 cm area to the right of the zone, and the gripper was initialized at top, middle, and bottom positions to the right of the block initialization area. This task is challenging because the robot often had to make point contact, which required maintaining the pushing direction while balancing the block.

d) **Plane**: In this task, the robot attaches the wheels into a toy plane by inserting both the peg and the rectangular base of the wheels, which can only be achieved with precise alignment. The plane was initialized at the four corners of a rectangular region 5 x 8 cm in size and the wheels at the corners of a 7 x 9 cm region with varying orientations. A successful trial means that the plane wheels are fully inserted and cannot be moved.

e) **Cube**: This task required the robot to insert a toy block into the corresponding slot on the shape sorting cube. During demonstrations, the cube was initialized at two positions 14 cm apart and the block could start from three positions forming a triangle of side 12 cm in length, in total making up 6 training initial states. At test time, we additionally placed the cube and the block at the midpoints of two adjacent initial positions.

f) **Nail**: In this task, the robot must align the claws of a toy hammer underneath the head of a nail. The nail was placed at a fixed position during demonstrations and the gripper was initialized to three poses as shown in Fig. 7a. At test time, the hammer was also reset to the midpoints, in a fashion similar to that of the cube task. Success means that upon the end of the trial, if we manually lift the gripper the nail goes off.

g) **Grasp-and-Place**: The robot must first grasp a toy apple, slightly lift it, and place it onto a paper plate. The apple was initially placed within the reachable area of the robot, whereas the position of the plate is fixed. The task is challenging because it requires multiple steps and if the apple is not lifted correctly, the plate will shift during placing. A success is deemed if the gripper is open in the end and the apple is placed within the plate.

h) **Grasp-Drop-Push**: This task aims to mimic the robot serving food. The robot needs to first grasp and lift a toy apple, drop it into a bowl, and push the bowl to be alongside a cup. The apple is randomly placed in a 20 x 30 cm area, the bowl could start anywhere in a 20 x 20 cm area, and the cup remains in place. This task is challenging because the sequence is long running, with several distinct actions. In addition, it requires 3D awareness to gently drop the apple into the bowl and reposition to push it without catching on the edge of the bowl. A success requires the complete execution of the whole sequence.

*i) **Grasp-Place-x2***: As an extension to the single object grasp-and-place, this task requires the robot to reach for, grasp, and carry a toy orange to a fixed point on the table and then, without stopping, move a toy apple to a different fixed point. Though the positions of the fruits were not varied, this is still challenging because it required long duration to complete and the round fruits easily roll. For a trial to be considered successful, the robot must set the fruits at their target positions smoothly without pausing.

*j) **Cloth*** Here, the robot must reach for a disheveled cloth on the table, grasp it, and lift it up into the air. During training, the cloth was placed anywhere within one of two 50 x 50 cm regions on the table, with the two regions 20 cm apart. During testing, the cloth was additionally placed between the two regions, in an unseen location. This task is made challenging by the fact that the cloth can appear in a visually diverse range of shapes and be piled to different heights. Success requires the robot to firmly grasp the cloth and lift it above the table.

*B. Loss Functions*

For all experiments, we used the same weighting coefficients of the loss functions $(\lambda_{l2}, \lambda_{l1}, \lambda_c, \lambda_{aux}) = (0.01, 1.0, 0.005, 0.0001)$, and we set $\lambda_g = 0.01$ for tasks involving gripper open/close. The vision networks $\theta_{vision}$ for all tasks were trained with an auxiliary loss to predict the current gripper pose $p_t \in \mathbb{R}^9$, represented by three points on the end effector, as well as another auxiliary loss to predict the gripper pose at the final time step $p_T$. For the plane and cube tasks, where the left gripper was used for holding an object, the vision networks were also trained with an auxiliary loss to predict the current left gripper pose. For the pushing, grasp-and-place, and grasp-drop-push tasks, an additional auxiliary loss for the vision networks was used to predict the current object position, which was inferred from the full history of right gripper pose and open/close status. Note the labels for all auxiliary prediction tasks were only provided during training.

*C. Neural Network Policy*

We represent the control policies using neural networks with architecture described in Section IV-A. For all experiments, initial values of network parameters were uniformly sampled from [-0.01, 0.01], except for the filters in the first convolution layer for RGB images, which were initialized from GoogLeNet [52] trained on ImageNet classification. Policies were optimized using ADAM [53] with default learning rate of $0.001$ and batch size of $64$. Our hyperparameter or architecture search was limited to: a) number of fully-connected hidden layers following the CNN (either one layer of 100 units or two layers of 50 units), b) whether to feed back auxiliary predictions $s_t$ to the subsequent layer (see Eq. 3), and c) $l_2$ weight decay of $\{0, 0.0001\}$. Typically, the policy achieved satisfactory performance with under three variations from our base architecture.

ACKNOWLEDGMENT

We thank Yan (Rocky) Duan for constructive writing suggestions and Mengqiao Yu for valuable assistance with the supplementary video. This research was funded in part by the Darpa Simplex program, an ONR PECASE award, and the Berkeley Deep Drive consortium. Tianhao Zhang received support from an EECS department fellowship and a BAIR fellowship. Zoe McCarthy received support from an NSF Fellowship.